\documentclass[10pt,twocolumn,letterpaper]{article}

\usepackage[pagenumbers]{cvpr} 

\usepackage{graphicx}
\usepackage{amsmath}
\usepackage{amssymb}
\usepackage{booktabs}
\usepackage{nicefrac}
\usepackage{xcolor}
\usepackage{mathtools}
\usepackage{enumitem}

\usepackage{siunitx}
\usepackage{amsmath,amsfonts,bm,physics}

\def\eqref#1{equation~\ref{#1}}

\def\1{\bm{1}}

\def\vc{{\bm{c}}}

\def\vz{{\bm{z}}}

\DeclareMathAlphabet{\mathsfit}{\encodingdefault}{\sfdefault}{m}{sl}
\SetMathAlphabet{\mathsfit}{bold}{\encodingdefault}{\sfdefault}{bx}{n}

\def\gN{{\mathcal{N}}}

\DeclareMathOperator*{\argmin}{arg\,min}

\usepackage[pagebackref,breaklinks,colorlinks]{hyperref}

\newcommand{\revised}[1]{\textcolor{black}{#1}}

\usepackage[capitalize]{cleveref}
\crefname{section}{Sec.}{Secs.}
\Crefname{section}{Section}{Sections}
\Crefname{table}{Table}{Tables}
\crefname{table}{Tab.}{Tabs.}

\newcommand{\customfootnotetext}[2]{{
  \renewcommand{\thefootnote}{#1}
  \footnotetext[0]{#2}}}

\def\cvprPaperID{9409} 
\def\confName{CVPR}
\def\confYear{2023}

\newcommand{\name}{\textsc{3D-LDM}\xspace}

\begin{document}

\title{\name: Neural Implicit 3D Shape Generation with Latent Diffusion Models}
\author{Gimin Nam$^*$\\
National University of Singapore\\
\and
Mariem Khlifi$^*$\\
Technical University of Munich\\
\and
Andrew Rodriguez\\
Bernard M. Baruch College \\
\and
Alberto Tono\\
Stanford University \& Computational Design Institute \\
\and
Linqi Zhou\\
Stanford University\\
\and
Paul Guerrero\\
Adobe\\
}

\twocolumn[{%
\renewcommand\twocolumn[1]{#1}%
\maketitle
\vspace{-30pt}
\begin{center}
    \centering
    \includegraphics[width=\linewidth]{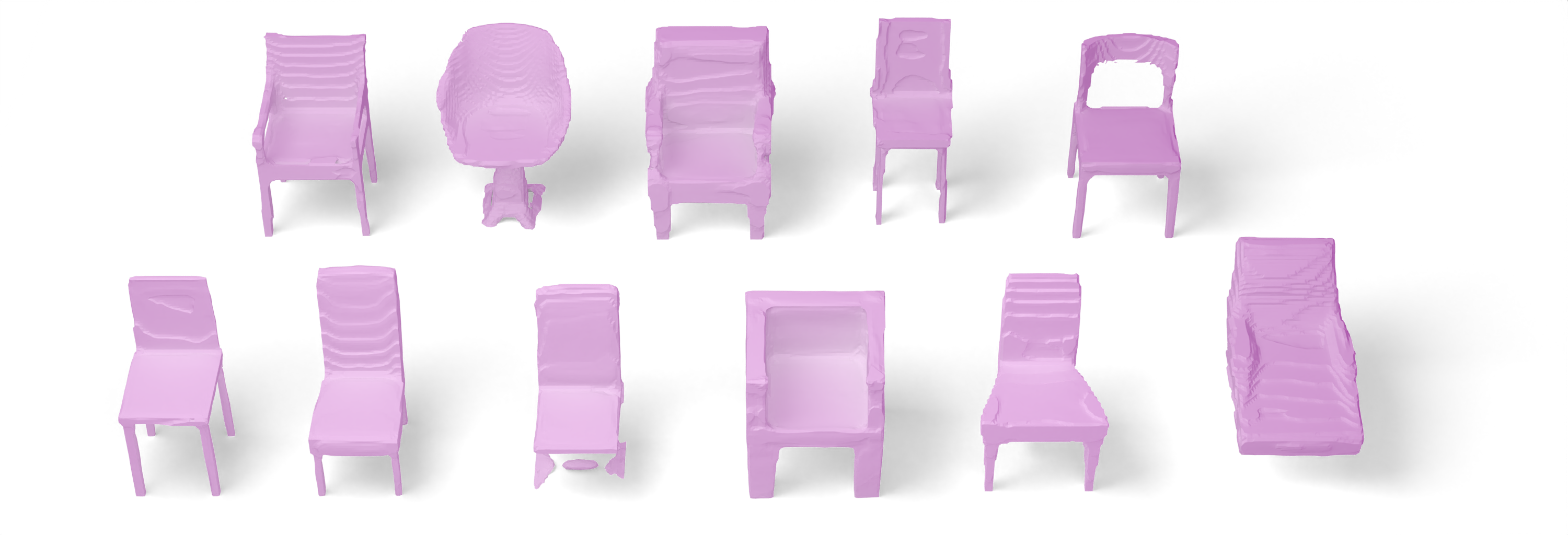}
    \vspace{-30pt}
    \captionof{figure}{We propose \name as a diffusion model that generates neural implicit representations of 3D shapes. Our approach applies a diffusion model to the latent space of an autodecoder for neural implicit shapes, allowing us to generate 3D shapes with continuous surfaces using a diffusion model.
 }
    \label{fig:teaser}
\end{center}%
}]

\customfootnotetext{$*$}{Joint first authors.}
\begin{abstract}
    Diffusion models have shown great promise for image generation, beating GANs in terms of generation diversity, with comparable image quality. However, their application to 3D shapes has been limited to point or voxel representations that can in practice not accurately represent a 3D surface. We propose a diffusion model for neural implicit representations of 3D shapes that operates in the latent space of an auto-decoder. This allows us to generate diverse and high quality 3D surfaces. We additionally show that we can condition our model on images or text to enable image-to-3D generation and text-to-3D generation using CLIP embeddings.
    Furthermore, adding noise to the latent codes of existing shapes allows us to explore shape variations.
\end{abstract}


\section{Introduction}
\label{sec:introduction}

Denoising Diffusion Probabilistic Models (DDPMs) \cite{DDPM} are an emerging paradigm for generating novel high-quality and diverse content, revolutionizing workflows to create images~\cite{nichol2021glide, aditya2022dallie2,saharia2022imagen,yu2022parti}, audio~\cite{kong2020diffwave, chen2020wavegrad}, and video~\cite{uriel2022makeavideo, ho2022imagenvideo}. DDPMs provide several advantages over other generative models like GANs: they tend to improve coverage of the data distribution, tend to be more stable at training time, and facilitate applications like inpainting.
Most DDPMs work on 2D images, leveraging their canonical representation as pixel grids. Only recently have researchers started adopting them to generate 3D content, which has a less standardized representation, ranging from voxels, point clouds, meshes, and implicit functions. This wide variety of possible representation makes an application of DDPMs to 3D shapes less straight-forward.

A large body of prior work on generating 3D shapes focuses on voxels \cite{brock2016generative, wu2016learning} and point clouds~\cite{li2018point,cai2020learning, yang2019pointflow, luo2021diffusion, pointvoxeldiffusion, zeng2022lion}. However, neither format is ideal for downstream creative needs: voxels are computationally and memory intensive and thus difficult to scale to high resolution; point clouds are light-weight and easy to process, but require a lossy post-processing step to obtain surfaces, which are required for some down-stream applications like rendering or modelling.
On the contrary, neural implicit functions~\cite{deepSDF , mildenhall2020nerf} do provide an accurate and efficient representation for surfaces. However, it is not immediately clear how we can generate such a representation with a diffusion model. 

In this work, we propose \name, a novel method for generating neural implicit representations of 3D shapes. We have been inspired by Latent Diffusion Models (LDMs) for 2D images~\cite{rombach2022high}.
Our key insight is that we can generate neural implicit representations of 3D shapes (Signed Distance Functions - SDFs \cite{yi2020coalesce, remelli2020meshsdf, zekun2020dualsdf}) by applying a simple diffusion model to the latent space of an auto-decoder for 3D shapes like DeepSDF~\cite{deepSDF}, therefore bypassing the need to directly operate on continuous surfaces. We show that this approach is significantly better than directly sampling from the latent space of the autodecoder and outperforms the previous state-of-the-art in unconditional 3D shape generation. Unlike existing diffusion-based approaches, we can generate neural implicit shape representations that allows us to extract high-quality surfaces. Compared to existing GAN-based models for generating neural implicit representations~\cite{ihgan, liu2022isfgan, monteiro2021pigan}, our diffusion-based approach achieves better data coverage and sample quality. Additionally, \name can easily be conditioned on context variables such as CLIP embeddings or images to enable both text-to-3D and image-to-3D generation. We also show that \name can create variations of existing shapes by adding small amounts of noise, enabling a form of guided shape exploration.

In summary, our contribution is threefold:
\begin{itemize}
    \setlength\itemsep{0pt}
    \item We propose a latent diffusion model that generates 3D implicit functions.
    \item We demonstrate text- and image-conditioned generation.
    \item We propose guided shape exploration based on adding noise to generated or existing shapes.
\end{itemize}

\section{Related Work}
\label{sec:related_work}

\name is a diffusion model for 3D shapes represented as neural implicit functions, thus we review the work on 3D shape generation, 3D diffusion models, and methods that generate neural implicit functions, also in \revised{a conditional setting.}

\textbf{3D deep generative models.} GAN \cite{achlioptas2018learning, marrnet, GET3D, valsesia2018learning, dongwook2019pointGAN}, Flow \cite{yang2019pointflow,SoftFlow,klokov20DPF-Net}, Score-Matching \cite{ShapeGF}, Autoregressive \cite{sun2020pointgrow}, Diffusion (DDPM) \cite{luo2021diffusion, pointvoxeldiffusion} or Energy (EBMs) \cite{SDEdit}  aim to approximate the distribution $p_\theta$ of a dataset $\mathcal{D}$. Thus, sampling from this data distribution allows the generation of novel shapes. However, since no ground truth for the new shapes is available, it requires likelihood-based evaluation metrics \cite{achlioptas2018learning}, to better evaluate if the model is overfitting on that distribution or \revised{generating novel shapes.} 

\textbf{3D diffusion models.} 
DDPMs \cite{DDPM, langevin15, DDIM, yang2022diffusion}, cover the current state-of-the-art in image generation tasks outperforming GANs  \cite{diffbetterthanGAN}. These models belong to latent variable models that have been inspired by non-equilibrium in thermodynamics literature and seek to gradually inject noise to the data distribution until an isotropic Gaussian is reached \cite{langevin15, bansal2022cold}. This paradigm has been successfully adopted in 3D domains \cite{luo2021diffusion, pointvoxeldiffusion, zeng2022lion,GET3D,hui2022neural}, showing promising results for diverse and high-quality shape generation. 
However, most works consider point clouds, or voxels, as the 3D representation, which are inherently limiting for graphics-based applications such as animation. 
 We instead consider implicit functions, \ie continuous surfaces, as our data representation which is more ideal for graphics-based rendering. Furthermore, previous diffusion models have been mainly applied to the data distribution, while we consider modeling latent distribution, echoing the recent success demonstrated by \cite{DDIM, zeng2022lion}. 

\textbf{3D shape generation via implicit function.}
 Neural implicit representations seek to parameterize surfaces with continuous functions, making them simple while preserving fine-grained details. Prior works \cite{IMnet, Occnet2019, deepSDF, zekun2020dualsdf, paritosh2022AutoSDF} adopted implicit representations for reconstruction tasks and some recent works such as Spaghetti \cite{hertz2022spaghetti} and Neural Wavelet \cite{hui2022neural} started using it for generation tasks. Neural Wavelet \cite{hui2022neural} and Spaghetti \cite{hertz2022spaghetti} respectively focused on multi-scale and part-aware implicit surface generation, which overcomes previous limitations of unstructured point cloud \cite{pointvoxeldiffusion, luo2021diffusion} and voxel-based methods \cite{pointvoxeldiffusion}. 
 Similarly, LION \cite{zeng2022lion} overcomes the above-mentioned limitations \revised{} via a differentiable Shape as Point (SAP) module \cite{Peng2021SAP}. Unfortunately, LION does not leverage SAP's differentiable property within its network, adding an extra computational burden. 

\textbf{Conditional Generation.} Generative models can be conditioned on additional contexts such as classes \cite{diffbetterthanGAN,ho2021classifierfree}, latent space \cite{vaesmeetdiff}, noise signal \cite{diffconditionnoise} or others \cite{ho2021cascaded, ho2021classifierfree}.
Conditioning allows for more control in the generative process. While previous approaches focused on 2D images, few examples successfully translated to 3D objects or scenes. PVD~\cite{pointvoxeldiffusion} \revised{conditions} on a depth-map. In concurrent work, GAUDI~\cite{gaudi} uses sparse images to generate novel 3D scenes, disentangling camera poses from scene representation\revised{. However, }novel 3D scene generation is not demonstrated conclusively. Our method can be conditioned on a CLIP-embedding that can be obtained from text or images, similar to previous work on 2D image generation~\cite{clipforge, clipnerf, khalid2022clipmesh, text2mesh, dreamfieldszeroshot, poole2022dreamfusion}.

\section{Method}
\label{sec:method}
\begin{figure*}[]
  \centering
\includegraphics[width=\linewidth]{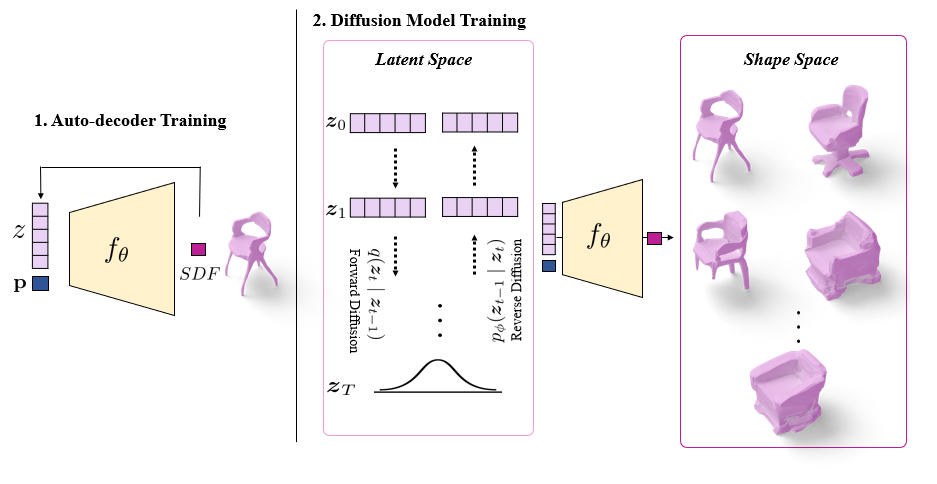}
  \caption{Overview of our architecture. In the first Stage, an auto-decoder is pre-trained to represent a set of 3D shapes; in the second stage, a diffusion model is trained on the latent space of the auto-decoder. At generation time, the diffusion model creates latent shape codes that are decoded into neural implicit 3D shapes by the pre-trained auto-decoder.}
  \label{fig:arch}
\end{figure*}

\name is a diffusion model for 3D shapes that is applied to a probabilistic auto-decoder \cite{Tan1995ReducingDD, deepSDF}.
We train it following a two-stage process. First, a latent space for a dataset of SDF-based shapes is constructed by training an auto-decoder that accepts a latent vector directly as an input. Second, the auto-decoder is frozen and a DDPM \cite{DDPM} is trained on the latent representation of the 3D shapes, effectively decoupling the 3D shape representation from the output representation used by the DDPM.
This allows us to use the DDPM to generate neural implicit 3D shapes when using an appropriate auto-decoder. An overview is shown in Figure~\ref{fig:arch}.
In the following sections, we describe the two stages in detail.

\subsection{Auto-decoder for Neural Implicit 3D Shapes}

\paragraph{Neural implicit shape representation.}
In an \emph{implicit representation}, the surface $S$ of a 3D shape is represented by the $0$-level set of a continuous function $f$ defined over the 3D space:

\begin{equation}
    S \coloneqq \{\mathbf{p}\ |\ f(\mathbf{p})=0\},
\end{equation}
with $\mathbf{p} \in \mathbb{R}^3$. We use the Signed Distance Function (SDF) as function $f$, which is defined as the signed distance to a (watertight) surface $S$; positive outside and negative inside.

In our approach, we work with \emph{neural implicit representations}, where the function $f$ is efficiently represented by a neural network $f_\theta$:
\begin{equation}
    S \coloneqq \{\mathbf{p}\ |\ f_\theta(\mathbf{p})=0\},
\end{equation}
with network parameters $\theta$. A \emph{set} of shapes $\mathcal{S}$ can be represented with $f_\theta$ by additionally conditioning $f_\theta$ on latent representations $\vz_i$ of the shapes:
\begin{equation}
    S_i \coloneqq \{\mathbf{p}\ |\ f_\theta(\mathbf{p}, \vz_i)=0\},
\end{equation}
with $S_i \in \mathcal{S}$.

We adopt the architecture proposed in DeepSDF~\cite{deepSDF} for $f_\theta$, with a few changes that empirically improve the quality of the generated shape distribution. We
refer readers to supplementary material for further details. 


\paragraph{Training setup.}
DeepSDF~\cite{deepSDF} has shown that $f_\theta$ can be trained efficiently in an encoder-less setup, called an \emph{auto-decoder}, where each shape $S_i$ is explicitly associated with a latent vector $\vz_i$, by storing a list of latent vectors $Z = (\vz_1, \dots, \vz_N)$, one for each of the $N$ shapes in $\mathcal{S}$. At training time, we optimize both $Z$ and $\theta$ to minimize the reconstruction error:
\begin{gather}
    \label{eq:loss}
    \argmin_{\theta, Z} \sum_{i=1}^N \mathcal{L}_\text{recon}(\vz_i, \text{SDF}_i) + \frac{1}{\lambda^2} \mathcal{L}_\text{reg}(\vz_i), \text{ with} \\
    \mathcal{L}_\text{recon} \coloneqq \mathbb{E}_{\mathbf{p} \sim \mathcal{P}} \left[\|f_\theta(\mathbf{p}, \vz_i) - \text{SDF}_i(\mathbf{p})\|_1\right], \text{ and} \nonumber \\
    \mathcal{L}_\text{reg} \coloneqq \|\vz_i\|_2^2.
\end{gather}
$\text{SDF}_i$ is the ground truth signed distance function for shape $S_i$, and $\mathcal{P}$ is the distribution of training sample points, which are either sampled uniformly in the shape bounding box, or close to the shape surface. The term $\mathcal{L}_\text{reg}$ regularizes the magnitude of the latent vectors to improve their layout in the latent space and make them easier to generate in our second stage; $\lambda$ is the regularization factor. Similar to DeepSDF, we clamp the predicted and ground truth signed distances to an interval $[-\delta, \delta]$, as predicting accurate distances far away from the surface is not necessary. \revised{For further details about the training setup, we refer the reader to Section \ref{sec:implementation_details} in the supplementary material.} 

Note that we cannot directly sample novel shapes from the latent space of the auto-decoder $f_\theta$, since the probability distribution of the latent data samples $\vz_i$ is unknown. Instead, we train a diffusion model on the samples $\vz_i$.

\subsection{Latent Diffusion Model}
\label{sec:latent_diffusion}

We base our diffusion model on DDPM~\cite{DDPM}. Diffusion models consist of two processes, a forward process and a reverse process.

\paragraph{Forward process.}
The forward process iteratively adds Gaussian noise to the shape latents in multiple steps, such that their distribution is indistinguishable from the standard normal distribution after $T$ steps. For each shape latent $\vz$,

we obtain a sequence of latent vectors $\vz^0,\dots,\vz^T$, each is created by adding noise to the previous vector in the sequence:

\begin{align}
    q\left(\vz^t \mid \vz^{t-1}\right) = \mathcal{N}\left(\vz^t ; \sqrt{1-\beta_t} \vz^{t-1}, \beta_t \mathbf{I}\right),
\end{align}
where $\vz^0 = \vz$ is the noiseless shape latent and $\mathcal{N}$ is the normal distribution. The scalars $\beta_t \in [0,1]$ form a variance schedule that defines the amount of noise added in each step. A closed-form solution for the distribution at any step $t$ is given by:
\begin{gather}
     q(\vz^t \mid \vz^0) = \mathcal{N}(\mathbf{x}_t; \sqrt{\bar{\alpha}_t}\mathbf{x}_{0}, (1-\bar{\alpha}_t) \mathbf{I}), \nonumber\\
     \text{with } \bar{\alpha}_t \coloneqq \prod_{s=1}^{t} \alpha_s \text{ and } \alpha_t \coloneqq (1-\beta_t). 
\end{gather}
The joint probability for the forward process is then:
\begin{equation}
q\left(\vz^{1: T} \mid \vz^0\right)=\prod_{t=1}^T q\left(\vz^t \mid \vz^{t-1}\right).
\end{equation}

\paragraph{Reverse process.}
The reverse process attempts to reverse each diffusion step and is defined as:
\begin{gather}
    \label{eq:reverse_step}
    p_\phi\left(\vz^{t-1} \mid \vz^t\right)=\mathcal{N}\left(\vz^{t-1} ; \mu_\phi\left(\vz^t, t\right), \sigma_t^2\mathbf{I}\right), \\
     \text{ where } \sigma_t^2 = \frac{1-\bar{\alpha}_{t-1}}{1-\bar{\alpha}_t}\beta_t. \nonumber
\end{gather}
Here, the mean $\mu_t$ at step $t$ is predicted by a neural network $\mu_\phi$ with parameters $\phi$.

Note that only the mean of the distribution needs to be predicted, as the variance can be computed in closed form from the variance schedule, please see DDPM~\cite{DDPM} for a derivation. The joint probability of the reverse process is defined as:
\begin{equation}
    p_\phi\left(\vz^{0: T}\right)=p\left(\vz^T\right) \prod_{t=1}^T p_\phi\left(\vz^{t-1} \mid \vz^t\right) \quad,
\end{equation}
where $p(\vz^T) = \gN(0, \mathbf{I})$. Thus, one can generate samples by first sampling from $p(\vz^T)$ and iteratively following the reverse process $p_\phi (\vz^{t-1} \mid \vz^t)$ until a final shape latent $\vz^0$ is produced.

\paragraph{Training setup.}
The network is trained by minimizing Kullback-Leibler divergence between the joint distributions $q(\vz^{1:T}\mid \vz^0)q(\vz^0)$ and $p_\phi\left(\vz^{0: T}\right)$.
Ho et al.~\cite{DDPM} note that this divergence can be minimized more stably and efficiently by training a network to predict the total noise $\vz^t - \vz^0$, of an input $\vz^t$ rather than the mean $\mu_t$ of a single denoising step:
\begin{gather}
    \argmin_{\phi} \sum_{i=1}^N \mathbb{E}\left[ \|\epsilon_\phi(\vz^t, t) - (\vz^t-\vz_i)\|_2^2 \right],\\
    \text{ with } t \sim \mathcal{U}(1, T) \text{ and } \vz^t \sim q(\vz^t|\vz_i). \nonumber
\end{gather}
$\mathcal{U}$ denotes the uniform distribution and $\epsilon_\phi$ is a network with parameters $\phi$. 
\revised{We refer readers to Section \ref{sec:implementation_details} in the supplementary material for further details about the implementation.} The mean $\mu_\phi$ for $p_\phi\left(\vz^{t-1} \mid \vz^t\right)$ is then defined as:
\begin{equation}
    \mu_\phi(\vz^t, t) = \frac{1}{\sqrt{\alpha_t}} \left(\vz^t - \frac{1-\alpha_t}{\sqrt{1-\bar{\alpha}_t}}\epsilon_\phi(\vz^t, t)\right).
\end{equation}

\subsection{Conditional Generation}\label{CG}
\begin{figure}[t]
    \centering
    \includegraphics[width=\linewidth]{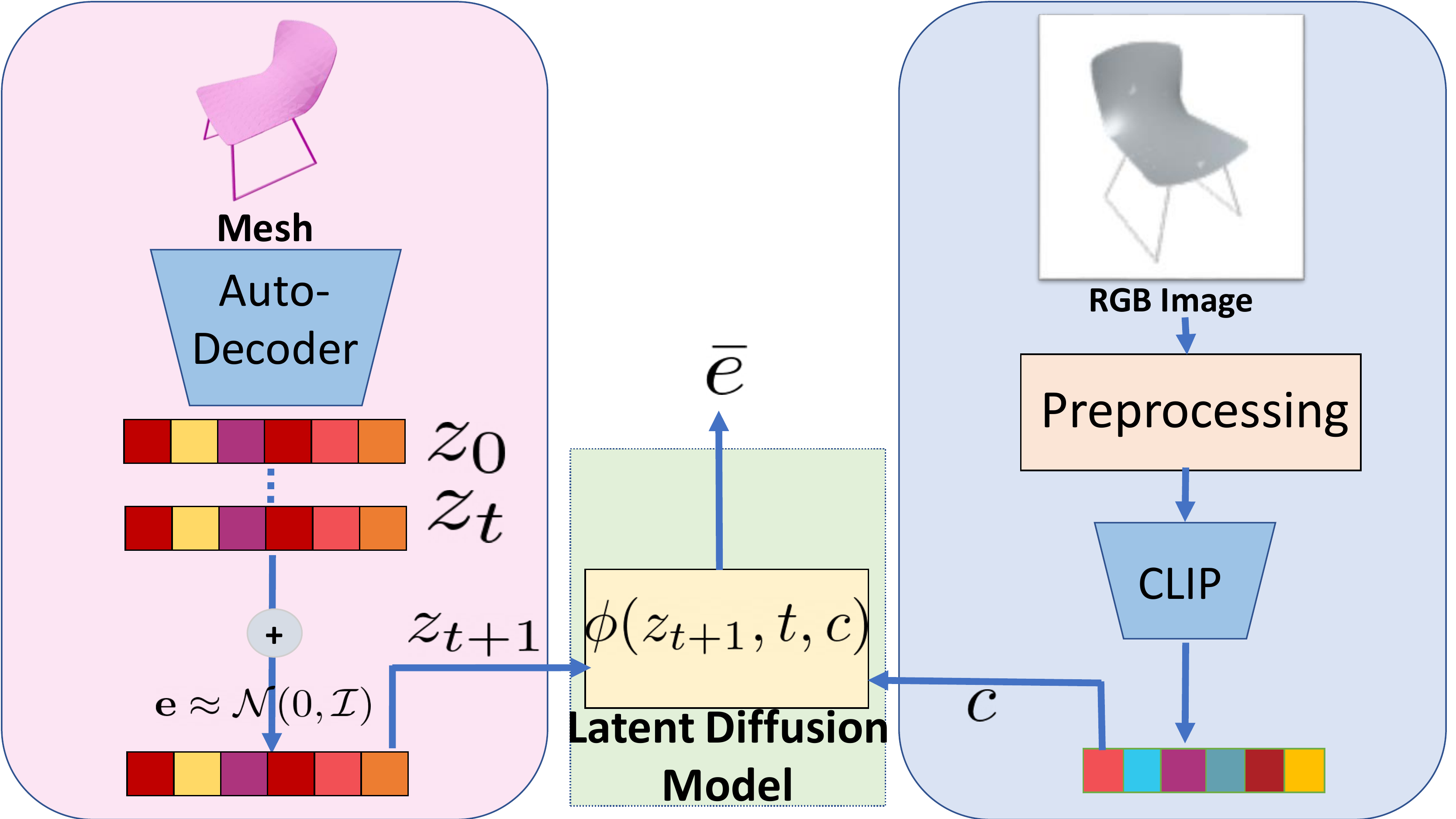}
    \captionof{figure}{CLIP-conditioned version of \name. We use a CLIP embedding of a given image or text prompt as additional input for our network $\epsilon_\phi$.}
    \label{fig:condition}
\end{figure}

Diffusion models can be conditioned on additional context variables. Information such as text and images can be encoded and the obtained embedding will be used by the diffusion model to guide the denoising process. We use a CLIP model~\cite{radford2021learning} trained on (image, text) pairs as a common embedding for text and images, \revised{enabling conditioning on both text and images. At training and generation time, we concatenate the CLIP embedding to the latent shape code as additional input to the network $\epsilon_\theta$}:
\begin{align}
    \mu_\phi(\vz^t, \vc, t) = \frac{1}{\sqrt{\alpha_t}} \left(\vz^t - \frac{1-\alpha_t}{\sqrt{1-\bar{\alpha}_t}}\epsilon_\phi(\vz^t, \vc, t)\right).
\end{align}
Figure~\ref{fig:condition} shows the architecture for conditional generation. 

\paragraph{Training setup.} \revised{We train on image embeddings only. Since CLIP has a shared text/image latent space, we can condition the trained model on text as well. Images for training are obtained by rendering each shape in our dataset from a random camera pose.}

\section{Results}
\label{sec:results}

We evaluate \name on unconditional generation and generation conditioned on text and images. Additionally, we show that \name can assist in shape exploration by proposing variations of a given shape.

\paragraph{Implementation and Training Details.}
Our latent shape vectors $\vz$ are $256$-dimensional and are randomly sampled from $\mathcal{N}(0,0.01)$. Similar to DeepSDF~\cite{deepSDF}, we are using an autodecoder with $8$ fully connected layers and $512$-dimensional hidden features. We set the regularization factor $\lambda = 100$. 
When generating samples, we run the inverse process for $30$k steps. 
We train the auto-decoder for $3000$ epochs ($\sim1$ day on a V100 GPU), and the diffusion model for $7000$ epochs ($\sim15$ hours on a V100 GPU).

\paragraph{Datasets.}
We evaluate our method on ShapeNet~\cite{shapenet2015}, a dataset containing synthetic 3D meshes of man-made objects. We use the \texttt{chairs} category ($6778$ shapes total, we randomly pick $6278$/$500$ as training/test set) and \texttt{airplanes} category ($4045$ shapes total, {$3545$}/{$500$} for training/testing).
For each shape, we pre-compute $500$k signed distance samples, using the sampling strategy and approximate inside/outside computation described in DeepSDF~\cite{deepSDF}. At training time, we evaluate our autodecoder on random subsets of $\sim16$k samples for each shape in a batch, picked to have a balanced number of positive and negative SDF values.

\begin{figure}[t]
  \centering
\includegraphics[width=\linewidth,height=0.5\linewidth]{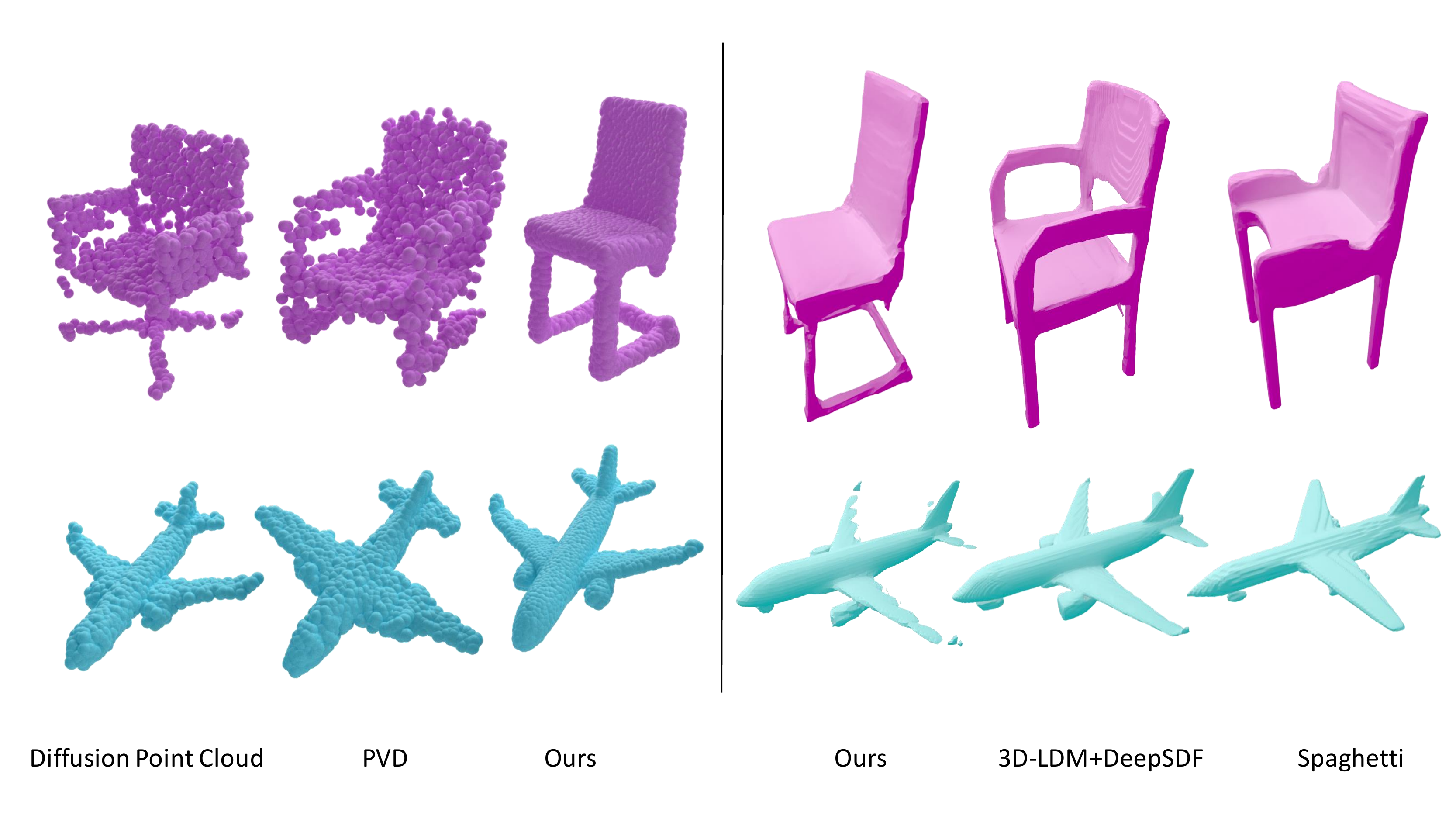}
  \caption{Qualitative comparison. Examples of generated shapes with \name (ours) compared to other baselines. We show both the surfaces and the point clouds from our method to compare to both baselines categories. Additional qualitative results are shown in Section \ref{sec:qual_comparison}.}
  \label{fig:comp}
\end{figure}

\subsection{Unconditional Generation}
We generate shapes unconditionally by starting from a noisy latent shape vector $\vz^T \sim \mathcal{N}(0, \mathbf{I})$ and denoising it using the reverse process described in Section~\ref{sec:latent_diffusion}. The denoised latent vector is decoded with our autodecoder into an SDF that can be converted into a meshed surface $S$ with Marching Cubes~\cite{Lorensen:1987:MarchingCubes}.

\begin{table*}[t]

\centering

\caption{\small Quantitative comparison of unconditional generation. We compare to three baselines (top group) and three ablations (bottom group). We achieve performance close to PointDiffusion, while generating neural implicit representations that can accurately represent surfaces instead of point clouds, and outperform the other baselines.
We multiply MMD by $10^{3}$ and the EMD by $10^{2}$.\\}
\label{tab:quant_comparison}
\scalebox{0.86}{
\setlength{\tabcolsep}{3pt}                     
\begin{tabular}{lcccccccccccc}                        
\toprule    
&\multicolumn{6}{c}{Chairs}
&\multicolumn{6}{c}{Airplanes}\\
\cmidrule(lr){2-7}\cmidrule(lr){8-13}
&\multicolumn{2}{c}{MMD$\downarrow$}
&\multicolumn{2}{c}{COV(\%)$\uparrow$}
&\multicolumn{2}{c}{1-NNA(\%)$\downarrow$}
&\multicolumn{2}{c}{MMD$\downarrow$}
&\multicolumn{2}{c}{COV(\%)$\uparrow$}
&\multicolumn{2}{c}{1-NNA(\%)$\downarrow$}
\\ 
\cmidrule(lr){2-3}\cmidrule(lr){4-5}\cmidrule(lr){6-7}\cmidrule(lr){8-9}\cmidrule(lr){10-11}\cmidrule(lr){12-13}
&CD     & \multicolumn{1}{c}{EMD}&CD                  &\multicolumn{1}{c}{EMD}&CD      &\multicolumn{1}{c}{EMD}
&CD     & \multicolumn{1}{c}{EMD}&CD                  &\multicolumn{1}{c}{EMD}&CD      &\multicolumn{1}{c}{EMD}
\\   \midrule
Point Diffusion~\cite{luo2021diffusion}             
&\textbf{13.6} & \textbf{7.52} & \textbf{47.8} & \textbf{46.0 } & \underline{60.0} & 69.9
&\textbf{3.3} & \textbf{3.33} & \textbf{48.4} & \textbf{48.2} & \textbf{66.5} & \textbf{70.9}
\\ 
Spaghetti~\cite{hertz2022spaghetti}             
&18.6 & 8.7 & 42.4 & \underline{43.8} & 66.99 & 72.4
&4.4 & 3.74 & 44.07 & 41.0 & 78.76 & 81.17
\\ 
PVD~\cite{pointvoxeldiffusion}             
&18.3 & 9.1 & 33.0 & 33.6 & 74.6 & 78.7
&4.0 & 3.61 & 36.2 & 37.2 & 88.2 & 93.39
\\ 
\midrule
DeepSDF\cite{deepSDF}                 
&44.9 & 45.0 & 12.0   & 11.6    & 94.6    & 94.4
&8.4&5.35&21.6&18&95.1&96.1
\\         
Our AutoDecoder
&59.8 & 19.96  & 7.4    & 6.6    & 95.3    & 95.4
&7.8&5.26&20.2&14&94.6&96.8
\\ 	
\name+DeepSDF        
&\underline{15.8 } & 8.27  & \underline{43.4 } & 42.6 & 61.7  & \underline{67.59} 
&\underline{3.9}&3.61&39.46&37.6&78.8&82.3
\\ 
\midrule
\name\textit{ (ours) }         
&16.8 & \underline{8.2} & 42.6 & 42.2 & \textbf{58.9} & \textbf{65.3}		 
&4.0&\underline{3.52}&\underline{46.2}&\underline{42.6}&\underline{74}&\underline{80.1}
\\ 
\bottomrule                                     
\end{tabular}
}
\end{table*}

\paragraph{Baselines.}
We compare our results to three state-of-the-art 3D shape generators: Point Diffusion~\cite{luo2021diffusion} directly applies the forward and reverse diffusion process to point clouds; PVD~\cite{pointvoxeldiffusion} applies a diffusion model to a hybrid point-voxel representation; and SPAGHETTI~\cite{hertz2022spaghetti} uses a part-aware GAN on neural implicit shapes. Additionally, we compare to three ablations of our method: sampling from a Gaussian fitted to the latent shape vectors $\vz_i$, sampling from a Gaussian fitted to the latent vectors of DeepSDF~\cite{deepSDF}, and using our method with the original DeepSDF as autodecoder instead of our slightly modified version. We use pre-trained versions of SPAGHETTI and PVD that were trained on the same dataset and categories. We use the test set provided in their code to compute metrics, taking care to have the same shape scaling to get comparable results. We re-train Point Diffusion and the ablations on our dataset.

\paragraph{Metrics.}
To quantitatively evaluate how similar a distribution of generated shapes is to the dataset distribution, we generate a set of $500$ shapes $\mathcal{S}_g$ and compare it to a set $\mathcal{S}_t$ of $500$ shapes from the test set, using metrics introduced in previous works on 3D shape generation~\cite{hertz2022spaghetti, yang2019pointflow, gal2021mrgan}:

\begin{itemize}[leftmargin=*, label={}]
    \item \textit{Minimum Matching Distance (MMD)}: Measures how far test samples are from generated samples:
    \begin{equation}
    \operatorname{MMD}\left(\mathcal{S}_g, \mathcal{S}_t\right) \coloneqq \frac{1}{\left|S_t\right|} \sum_{Y \in S_t} \min _{X \in S_g} D(X, Y),
    \end{equation}
    where $D(x,y)$ is a shape distance we will define below.
    \item \textit{Coverage (COV)}:
    Measures what percentage of test samples are covered by generated samples. A test sample is covered if it is the nearest neighbor of a generated sample:
    \begin{equation}
        \operatorname{COV}\left(\mathcal{S}_g, \mathcal{S}_t\right) \coloneqq 100\ \frac{|\mathcal{S}_t \cap \{\operatorname{NN}^D_{\mathcal{S}_t \cup \mathcal{S}_g}(S) \mid S \in \mathcal{S}_g\}|}{|\mathcal{S}_t|} ,
    \end{equation}
    where $\operatorname{NN}^D_\mathcal{X}(Y)$ is the nearest neighbor of $Y$ in $\mathcal{X}$ according to distance metric $D$.
    \item \textit{Nearest Neighbour (1-NNA)} Measures how well test samples and generated samples are mixed by penalizing samples from the test set and generated set that have their nearest neighbor in the same set.
\end{itemize}
\revised{We show results for the Chamfer Distance (CD)~\cite{Barrow:1977:Chamfer,fan2017point} and the Earth Mover's distance (EMD)~\cite{rubner1998metric}} as metric $D$. Both operate on {$2048$} uniform point samples from the shape surfaces $S$.

\begin{table}[t]
\centering
\caption{\small Time Complexity of the generation process per shape}
\label{tab:complexity}
\scalebox{0.86}{
\setlength{\tabcolsep}{3pt}                     
\begin{tabular}{ccccc}                        
\toprule  
 & Point Diffusion\cite{luo2021diffusion} & PVD\cite{pointvoxeldiffusion} & \name \textit{(ours)} \\
\midrule
Time (s) & \textbf{0.073 }& 62.05& 53.3 \\
 \bottomrule   
\end{tabular}
}
\end{table}

\paragraph{Results.}
\revised{Table~\ref{tab:quant_comparison} shows a quantitative comparison to the baselines. PointDiffusion and PVD generate point clouds as shape representation, while SPAGHETTI and \name create neural implicit shapes that allow for an accurate representation of the surface. We achieve performance close to PointDiffusion, while also having an accurate surface representation, and improve upon both PVD and SPAGHETTI. In the ablations, we can see that using a diffusion model to generate shape latent codes is significantly better than sampling them from a Gaussian fitted to the shape codes in the training set (DeepSDF and our AutoDecoder). We can also see that our architectural changes of the autodecoder result in a slight performance improvement over directly using DeepSDF as autodecoder (\name + DeepSDF).}
 
In Figure \ref{fig:comp}, we showcase examples of generated instances for \emph{chair} and \emph{airplane} categories. Additional qualitative examples are provided in Section \ref{sec:qual_comparison} of the supplementary material. We also provide results that show that our model does not overfit to the training data by comparing the generated shapes to their closest match in the training distribution. Further details are provided in Section \ref{sec:overfitting} of the supplementary material.

\revised{\paragraph{Computational Complexity.}
Table~\ref{tab:complexity} compares the computational complexity, i.e. the time required for a single shape synthesis, of our method to several baselines\cite{luo2021diffusion,pointvoxeldiffusion}.}

\subsection{Conditional Generation}

\begin{figure}[t]
  \centering
\includegraphics[width=\linewidth]{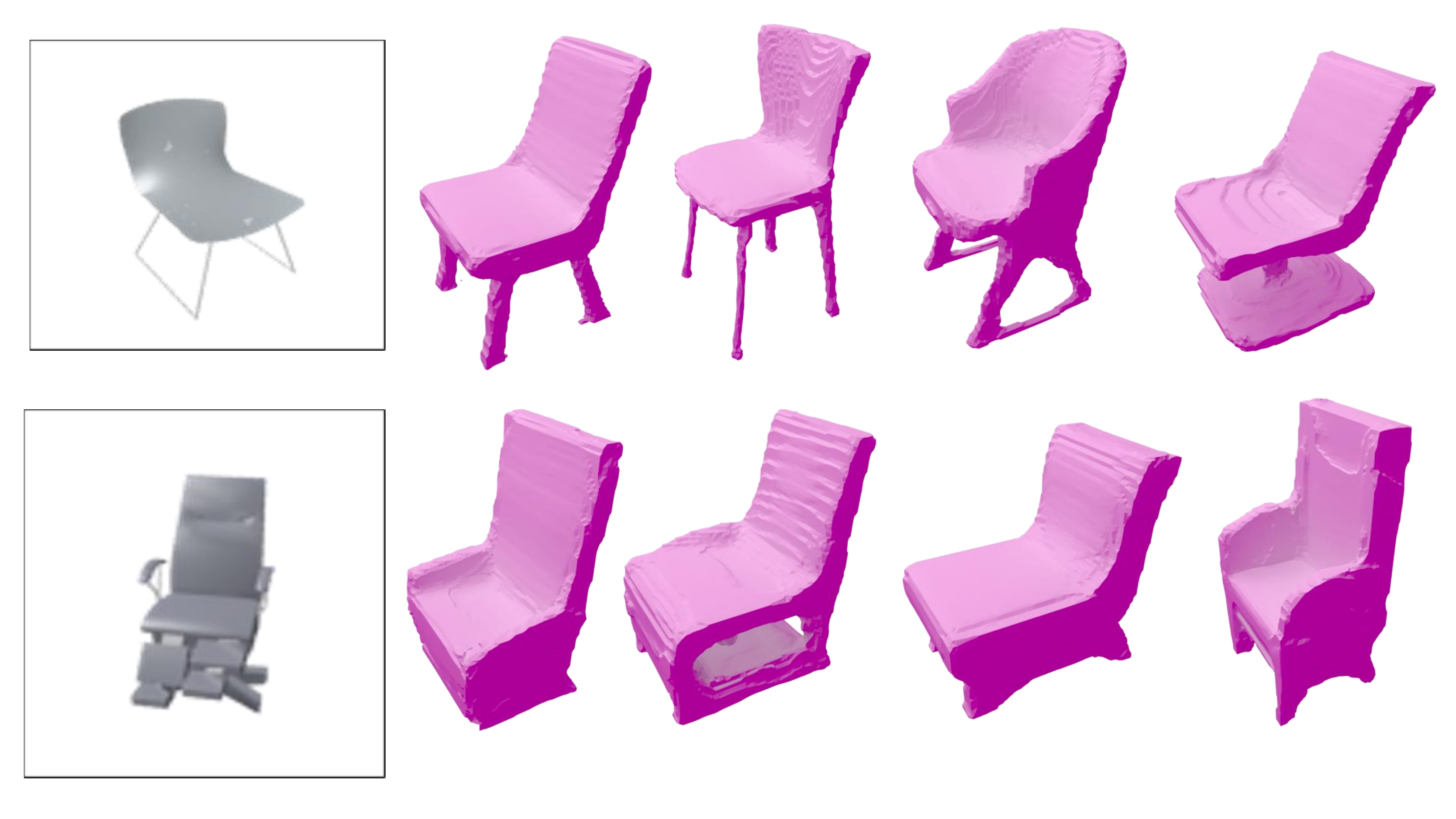}
  \caption{Image-to-Shape. \name conditioned on CLIP embeddings can generate shapes similar to an input image. In each row, we showcase multiple shapes generated by our model for two different image renderings of chairs extracted from the \textit{chairs} dataset.}
  \label{fig:imclip}
\end{figure}

\revised{We additionally evaluate our diffusion model conditioned on CLIP embeddings. The goal is to learn how to retrieve the best matching shapes given an input image or text prompt. We train with images of ShapeNet chairs rendered from random camera poses, as in \cite{choy20163dr2n2}, encoded with CLIP\cite{radford2021learning}. We showcase the obtained 3D shapes in \revised{Fig~\ref{fig:imclip}}, for each input image, we generate multiple shapes by using different random samples for the initial noise $\vz^T$.}

\begin{figure}[t]
  \centering
\includegraphics[width=\linewidth]{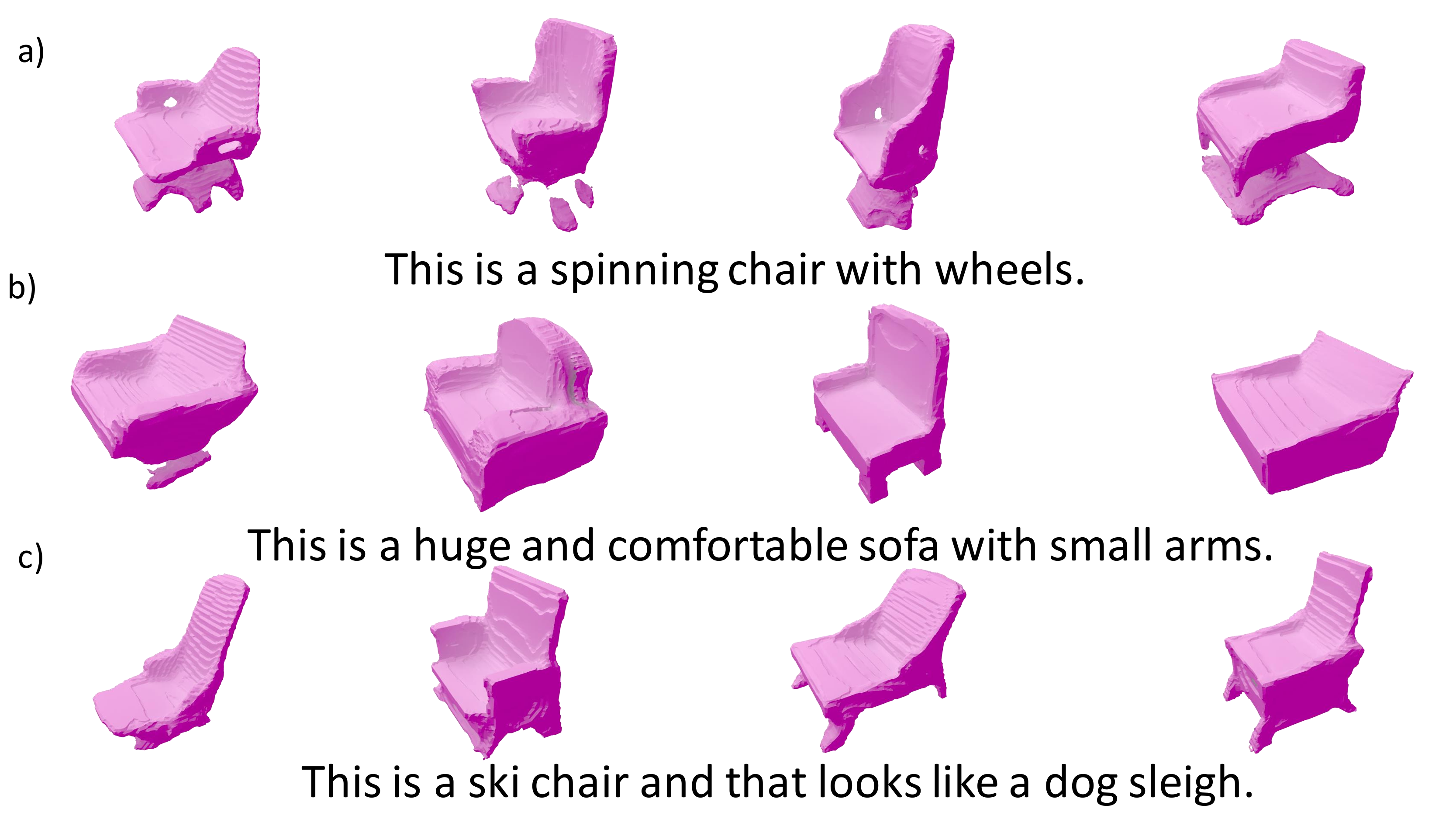}
  \caption{Text-to-Shape. \name conditioned on CLIP embeddings can generate shapes that match the description given by the text prompt. In each row, we showcase multiple shapes generated by our model for each text prompt.}
  \label{fig:textclip}
\end{figure}

We further repeat the experiment on text prompts and test the capability of our model to generate shapes matching the prompt. \revised{Note that the model does not need to be re-trained for this task, we use the same model that was trained with encoded images.} While some works have focused on category inspired prompts having the form "This is a sofa" or "this is a desk chair", we have evaluated our model against more descriptive text prompts. We can see that our model conditioned on CLIP yields shapes obeying the description while varying in overall appearance. We present a few examples in Figure \revised{\ref{fig:textclip}} showing chairs with wheels or chairs that look like a bed. In the second example, we can see that the chairs are larger and more comfortable. 

\revised{}

\subsection{Guided Shape Exploration}
\label{sec:guided_exploration}

\revised{A fundamental variable of a Diffusion Model is the maximum number of time-steps during the noising/denoising process. The number of time-steps $T$ used to transform the distribution of latent shape vectors $z^0$ to the standard normal distribution needs to be pre-defined.}

\revised{Generation of new shapes from scratch typically starts from pure noise at $t=T$. Given an initial latent code, however, we can explore variations of the corresponding shape by adding moderate amounts of noise $t<T$. The amount of added noise defines how similar the generated shape will be to the given initial shape. For $t \approx \mathcal{T}$, we obtain a randomly generated shape unrelated to the original whereas for $t<<\mathcal{T}$, the generated shape is identical to the initial shape. By choosing t to be moderately high, we obtain variations of the original shape where the differences amount to local styling properties.}

\begin{figure}[t]
  \centering
\includegraphics[width=\linewidth,height=0.5\linewidth]{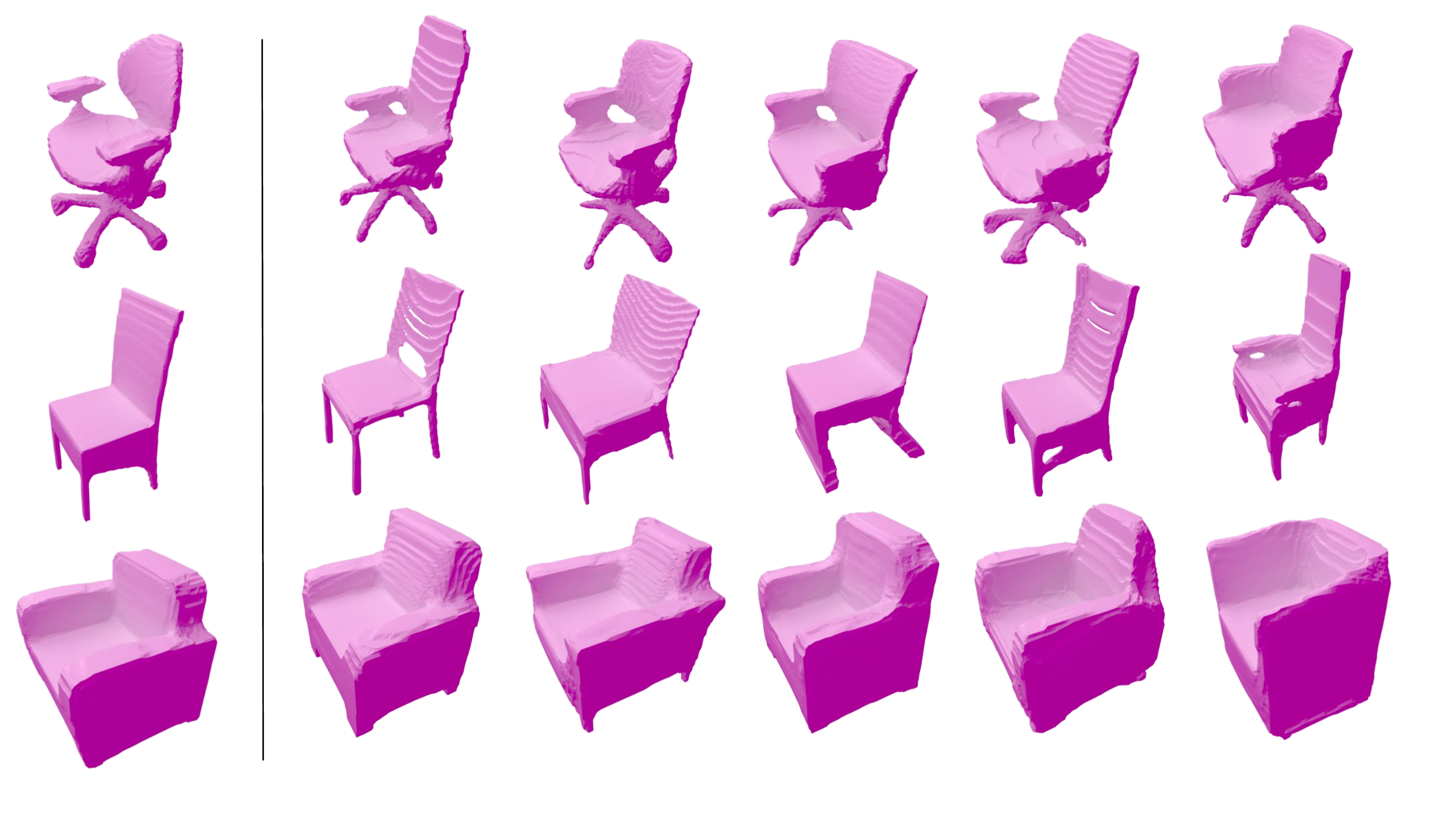}
  \caption{Guided shape exploration. By adding noise to the input shape (left column), we can generate variations of the input shapes that can be used for shape exploration. In each column, we show a different generated sample after adding 2000 noise steps to the input shape. We can see that the coarse shape of the input is preserved, but each variation generates different shape detail.}
  \label{fig:exploration}
\end{figure}

\revised{We conduct an experiment with 2000 steps of added noise and show examples in Figure \ref{fig:exploration}.}

\revised{}

\section{Conclusions}
\label{sec:conclusions}

We have introduced \name, a deep generative latent diffusion model for 3D shape generation for neural implicit functions. \name captures the distribution of the implicit representation of the SDF in a two-stage training. First, we trained the DeepSDF auto-decoder and then the Latent Diffusion Model on the latent space of the auto-decoder to ensure better generalization and fidelity for 3D content generation. As a result, \name ensures a performance on par with point cloud based diffusion models. Additionally, our model can be conditioned on text and images, and enables shape guided exploration by carefully adding and removing noise from the shapes. 

In the future, several interesting directions could be explored. First, we could jointly train on multiple categories instead of a single category to exploit the more accurate coverage of the data distribution that a diffusion model can provide over a traditional GAN. Second, we could generate textured scenes by adding a color output in addition to the signed distance. Finally, exploring NeRF generation by adding a differentiable volume renderer to our pipeline and training with multi-view images as supervision seems like an interesting direction.

{\small
\bibliographystyle{ieee_fullname}
\bibliography{ms}
}


\appendix

\section{Overview}
\label{sec:overview}

In the following, we provide additional architecture and implementation details (Section~\ref{sec:implementation_details}),
provide a qualitative comparison between \name and existing methods (Section~\ref{sec:qual_comparison}), evaluate the novelty of generated shapes by showing the nearest neighbors for a few generated shapes (Section~\ref{sec:overfitting}) and provide ablations for the choice of hyperparameters for our diffusion model (Section~\ref{sec:ablations}).
We will publish both the code and data we used for training.

\section{Implementation Details}
\label{sec:implementation_details}

\paragraph{Autodecoder Architecture}
We follow the architecture of DeepSDF~\cite{deepSDF} for our autodecoder, with a few changes. Like DeepSDF, we use an MLP with $8$ fully-connected layers, a $512$-dimensional feature space for hidden layers, and a skip connection from the input to the fourth hidden layer. Unlike DeepSDF, we do not use Dropout~\cite{srivastava2014dropout} or weight normalization~\cite{salimans2016weight}, and remove the tanh activation from the final layer, directly using the linear output instead. Empirically, we observed that this improves the quality and diversity of our generated shapes.

\paragraph{Denoiser Architecture}
The denoiser uses an MLP consisting of $8$ linear layers with softplus activations, $512$-dimensional hidden features, and three skip connections that concatenate the output of the first hidden layer to the input of hidden layers number $3$, $5$ and $7$. Additionally, each layer is conditioned on an embedding of the denoising step $t$. Similar to DDPM~\cite{DDPM}, we use a sinusoidal embedding~\cite{vaswani2017attention} of the time step. Additionally, each layer of the MLP, except for the first layer, transforms the time step embedding independently using a linear layer with SiLU activation~\cite{hendrycks2016gelu} before adding it to the layer input. Further details about the chosen hyperparameters for training the diffusion model are presented in Section \ref{sec:ablations}.

\paragraph{Additional Autodecoder Training Details}
To train our autodecoder, we use the training setup of DeepSDF~\cite{deepSDF}. We minimize the loss in Eq. 4 of our main paper using the Adam optimizer. For the point distribution of sample points $\mathcal{P}$, we use the same approach as in DeepSDF and pre-compute $500$k sample points that we randomly choose at training time. From these $500$k points, $5\%$ are uniformly distributed in the bounding box of a shape, and the rest is sampled near exterior surfaces of the shape. Exterior surfaces are identified as surfaces that are visible from a set of viewpoints $\mathcal{V}$ in a unit sphere around the shape. Points are first sampled on the exterior surfaces as intersection points of rays in view frustums that originate at viewpoints $\mathcal{V}$ and point at the origin, and then perturbed with Gaussian noise. Half of the surface points are perturbed with variance $\num{2.5e-3}$, the other half with variance $\num{2.5e-4}$. At training time, we create batches by randomly sampling a subset of $16$k points per shape.

\begin{figure*}[t]
  \centering
\includegraphics[width=\linewidth,height=0.5\linewidth]{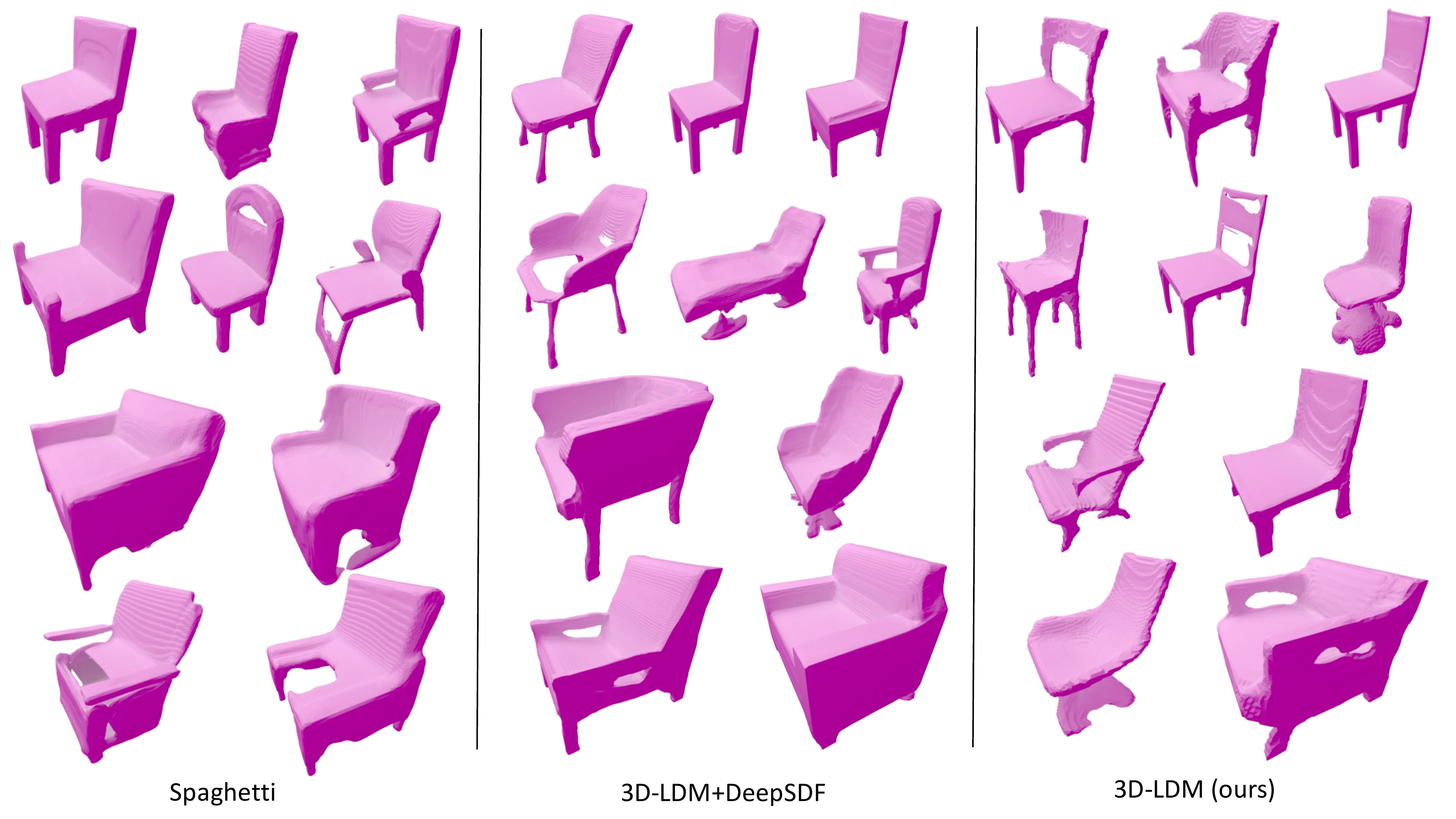}
  \caption{Qualitative comparison of generated shapes on ShapeNet Chairs. We compare $10$ randomly picked shapes generated by our method to $10$ shapes generated by each of the baselines returning surfaces similar to our method. For \name, \name+DeepSDF and SPAGHETTI~\cite{hertz2022spaghetti}, we visualize the 0-level set of the neural implicit representation (via a mesh extracted by Marching Cubes~\cite{Lorensen:1987:MarchingCubes}).
  }
  \label{fig:qual_comparison1}
\end{figure*}

\begin{figure*}[t]
  \centering
\includegraphics[width=\linewidth,height=0.5\linewidth]{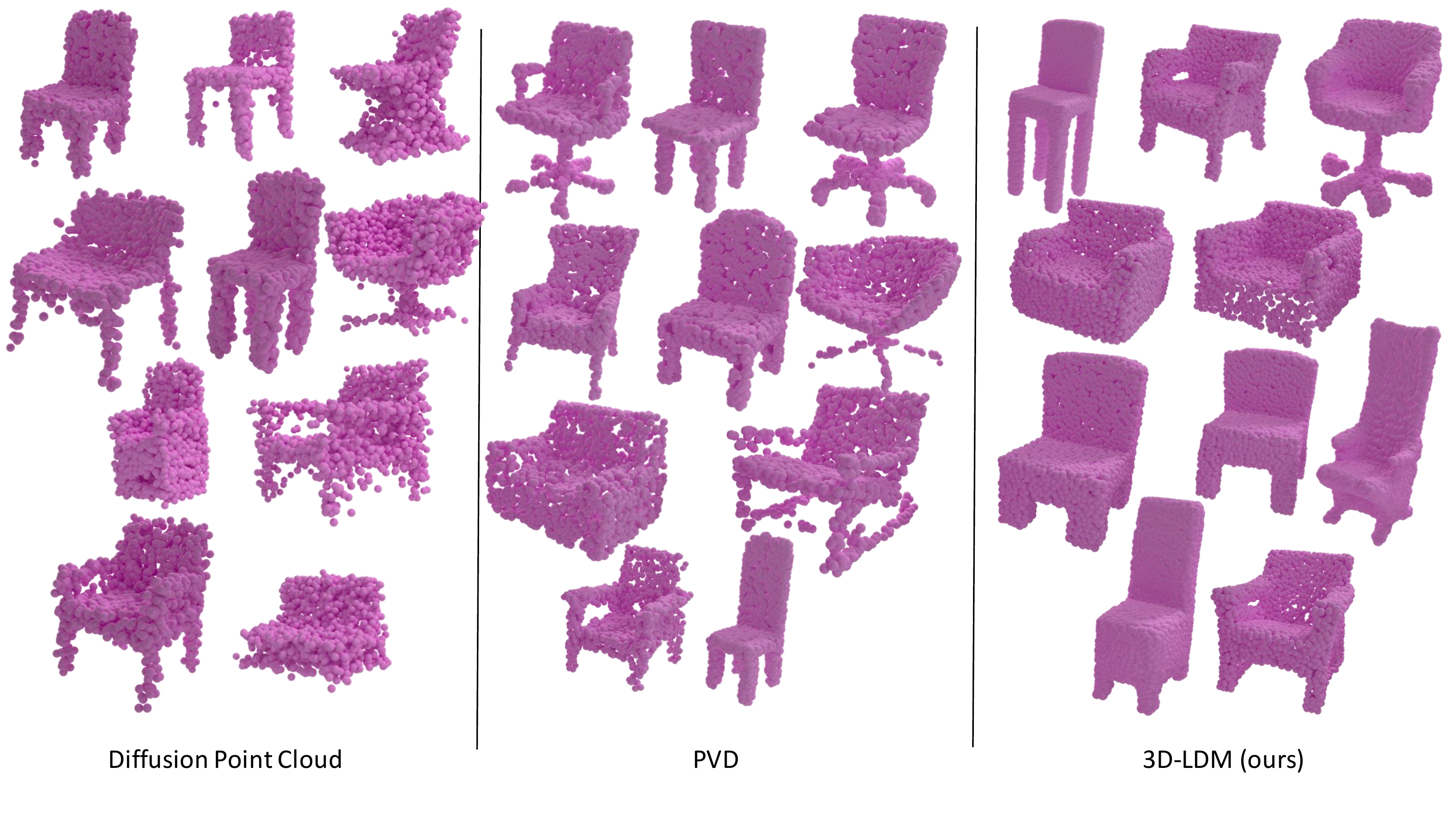}
  \caption{Qualitative comparison between 10 chairs generated by our model, picked at random from a subset of generated shapes, with point-cloud based methods from the baselines: Point Diffusion~\cite{pointclouddiffusion} and PVD~\cite{pointvoxeldiffusion}.}
  \label{fig:qual_comparison3}
\end{figure*}

\begin{figure*}[t]
  \centering
\includegraphics[width=\linewidth,height=0.5\linewidth]{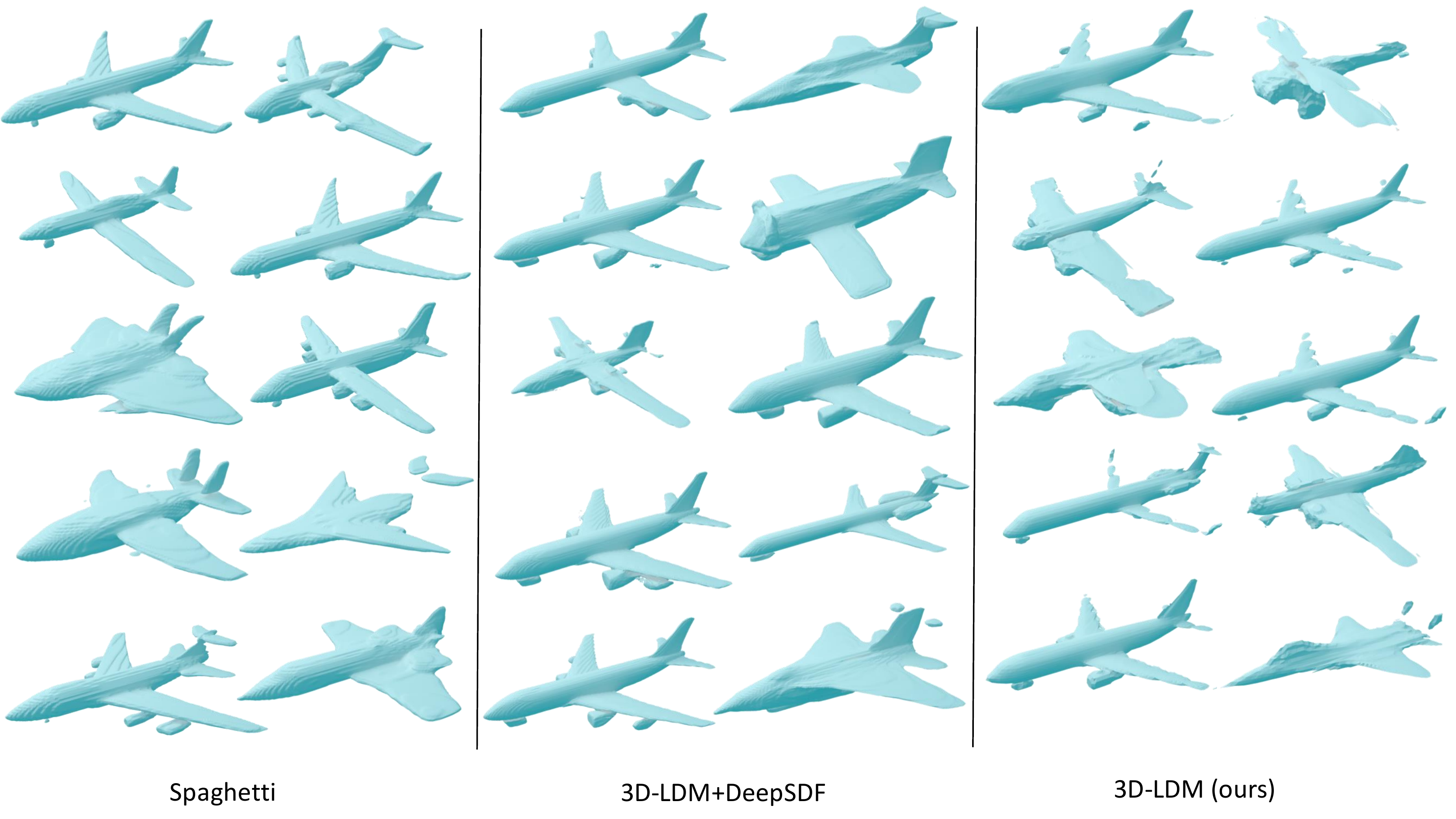}
  \caption{Qualitative comparison of generated shapes on ShapeNet Airplanes. We compare $10$ randomly picked shapes generated by our method to $10$ shapes generated by each of the baselines returning surfaces similar to our method. For \name, \name+DeepSDF and SPAGHETTI~\cite{hertz2022spaghetti}, we visualize the 0-level set of the neural implicit representation (via a mesh extracted by Marching Cubes~\cite{Lorensen:1987:MarchingCubes}).}
  \label{fig:qual_comparison2}
\end{figure*}

\begin{figure*}[t]
  \centering
\includegraphics[width=\linewidth,height=0.5\linewidth]{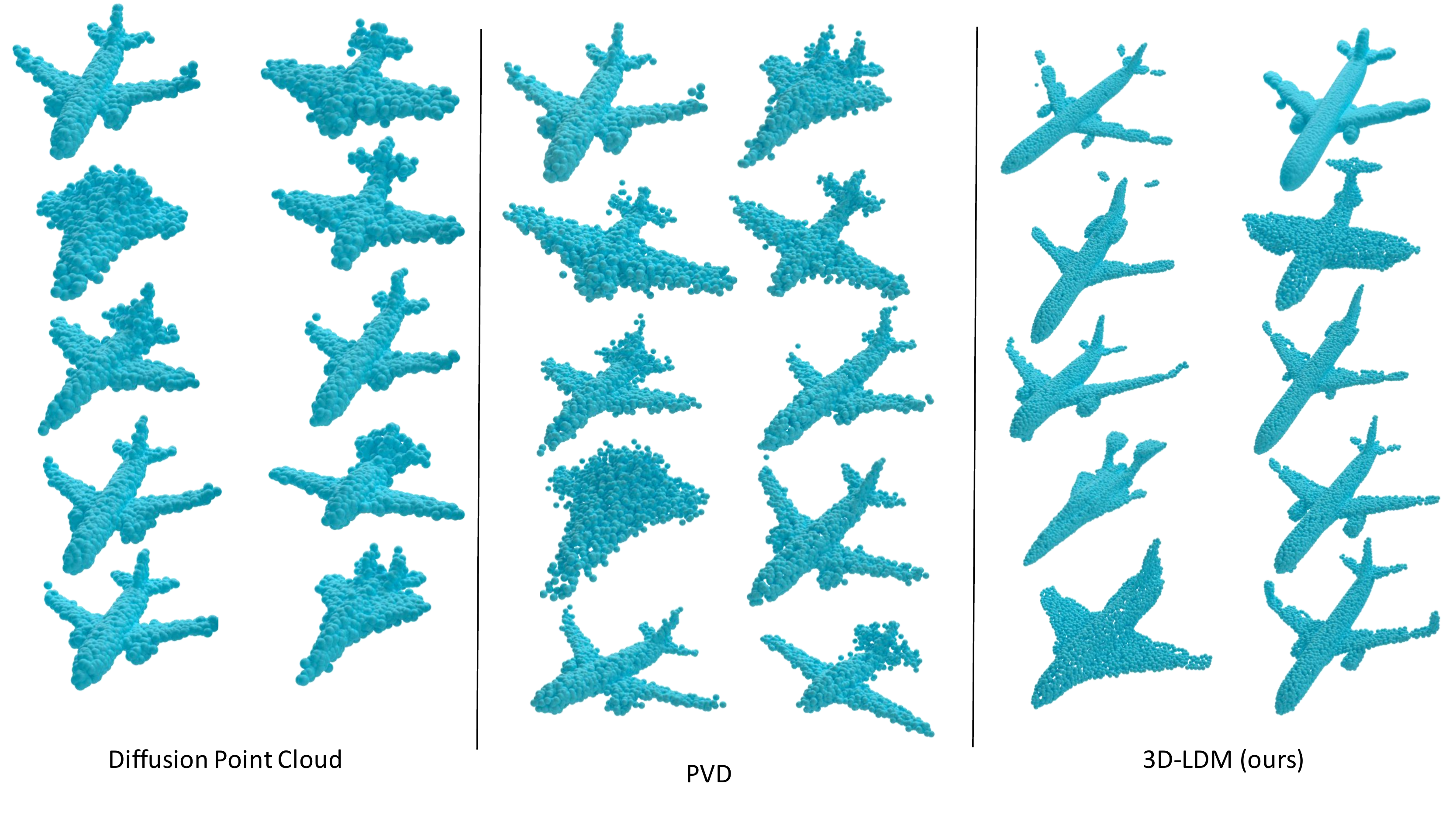}
  \caption{Qualitative comparison between 10 airplanes generated by our model, picked at random from a subset of generated shapes, with point-cloud based methods from the baselines: Point Diffusion~\cite{pointclouddiffusion} and PVD~\cite{pointvoxeldiffusion}.}
  \label{fig:qual_comparison4}
\end{figure*}

\section{Qualitative Comparison}
\label{sec:qual_comparison}

In Figures~\ref{fig:qual_comparison1}, ~\ref{fig:qual_comparison2}, ~\ref{fig:qual_comparison3} and ~\ref{fig:qual_comparison4}, we compare $10$ \emph{uncurated} shapes from our method and each of the baselines, on the Chairs and Airplanes categories, respectively. \\

Since PVD~\cite{pointvoxeldiffusion} and Point Diffusion~\cite{luo2021diffusion} create point clouds, we separately visualize their results compared to \name as point clouds in Figures~\ref{fig:qual_comparison3} and ~\ref{fig:qual_comparison4}. \\
For \name, \name+DeepSDF and SPAGHETTI, we visualize the surfaces reconstructed with Marching Cubes~\cite{Lorensen:1987:MarchingCubes} from the 0-level set of the neural implicit representation. We note that the shape quality is comparable across methods, with some artifacts in the results of each method. Recall that the quantitative results in Table~\ref{tab:quant_comparison} suggest that our generated distribution better covers the data distribution, i.e. the generated shapes better reflect the variety of shapes available in the data. While this is hard to show with a limited set of uncurated samples, we can see some indication of this behaviour in the chair examples, where our results also contain types of chairs that are less common in the dataset such as swivel chairs and deck chairs, in addition to the more common dining chairs and sofa chairs, while the examples from Spaghetti consist mainly of the more common chair types.
\\
While in the methods showcasing surfaces we detect discontinuities, point-cloud based methods usually remain noisy even after the denoising process is fully executed as we can clearly see in the generated shapes by PVD and Point Diffusion. However, these methods tend to have a superior synthesis capability than most methods which can be seen through the variety of shapes in Figures ~\ref{fig:qual_comparison3} and ~\ref{fig:qual_comparison4}. This synthesis quality is also valid for our approach \name which ensures good shape generation as supported by Table ~\ref{tab:quant_comparison} and additionally provides better distributed points with minimal outliers giving the shape a nicer appearance. Figures~\ref{fig:qual_comparison1}, ~\ref{fig:qual_comparison2}, ~\ref{fig:qual_comparison3}, ~\ref{fig:qual_comparison4} and Table ~\ref{tab:quant_comparison} show that with our method, we have a good variety of shapes, so overall it's on par with the popular diffusion-on-point-clouds methods, while also offering the possibility to directly visualize surfaces and get good quality of point distribution with minimal outliers.

\section{Shape Novelty}
\label{sec:overfitting}

\begin{figure*}[t]
  \centering
\includegraphics[width=\linewidth]{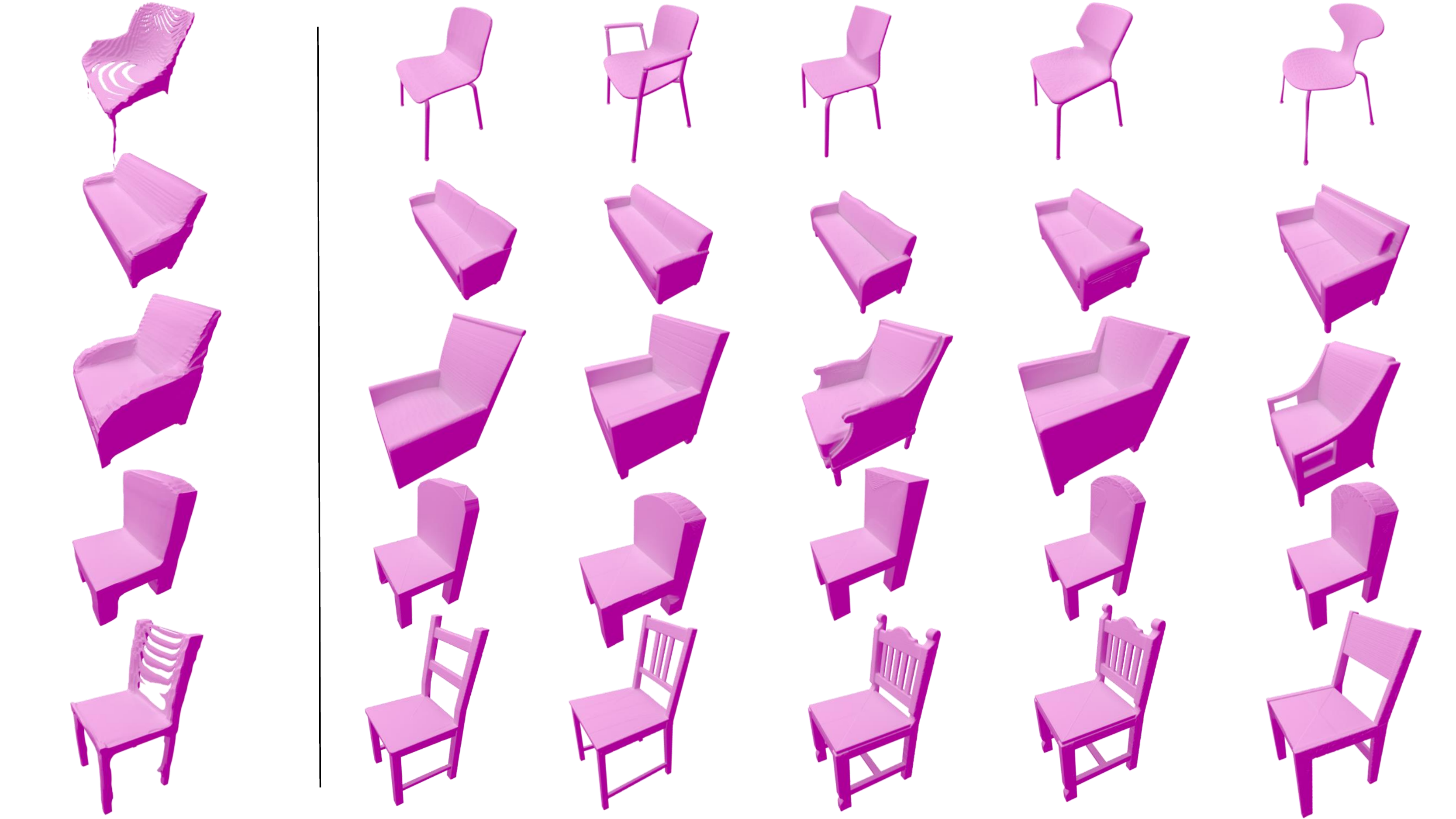}
  \caption{Novelty of generated shapes. We compare five randomly picked generated shapes to the five nearest neighbors in the training set, according to the Cosine Similarity. We can see that the generated shapes are significantly different from the training shapes, showing that our generated shapes are novel.}
  \label{fig:ovefitting}
\end{figure*}

In Figure~\ref{fig:ovefitting}, we evaluate the novelty of our generated shapes by comparing $5$ randomly chosen generated shapes (denoted by the latent codes $\mathbf{c_{i}}$) to their $5$ nearest neighbors (NN) in the training set (denoted by the latent codes $\mathbf{c_{j}}$) , using the Cosine Similarity as metric:
$$NN(\mathbf{c_{i}}) = \max_{j} cos(\mathbf{c_{i}},\mathbf{c_{j}})= \max_{j} \frac{<\mathbf{c_{i}},\mathbf{c_{j}}>}{||\mathbf{c_{i}}||.|||\mathbf{c_{j}}|}$$ 

The generated shapes have clear differences to their nearest neighbors in the training set, showing that the these generated shapes are indeed novel.

\begin{table*}[t]

\centering

\caption{\small Results of evaluation of different models for different hyperparameters. These metrics use the Chamfer Distance.
We multiply MMD by $10^{3}$.\\}
\label{tab:hyperparam_ablation}
\scalebox{0.86}{
\setlength{\tabcolsep}{3pt}                     
\begin{tabular}{lccccccc}                        
\toprule    
&\multicolumn{3}{c}{Hyperparameters}
&
&\multicolumn{3}{c}{Metrics}\\
\cmidrule(lr){2-4}\cmidrule(lr){6-8}
&Learning Rate & Batch Size & Timesteps & & MMD$\downarrow$ & COV$\uparrow$ & 1-NNA$\downarrow$
\\
\cmidrule(lr){2-2}\cmidrule(lr){3-3}\cmidrule(lr){4-4}\cmidrule(lr){6-6}\cmidrule(lr){7-7}\cmidrule(lr){8-8}
& 	1e-4 &	10 & 30000 & & 13.0 &	37.0 &	71.0\\

&\textbf{1e-5} &	\textbf{10} &	\textbf{30000} & &\textbf{11.3} &	\textbf{48.0} 	& \textbf{57.0}\\

& 	1e-6 &	10 & 30000 & & 13.8 &	30.0 &	73.5\\

& 	1e-4 &	24 & 30000 & & 12.7 & 39.0 &	66.0\\

& 	1e-5 &	24 & 30000 & & 12.4 	& 44.0 & 	61.5\\

& 	1e-5 &	24 & 10000 & & 30.9 &	9.0 &	93.5\\
\bottomrule                                     
\end{tabular}
}
\end{table*}

\section{Hyperparameter Ablations}
\label{sec:ablations}

In Table~\ref{tab:hyperparam_ablation}, we compare the performance of our approach under different hyperparameter settings. We show different settings for the learning rate, the batch size, and the total number of denoising steps.

Results show that $30$k steps significantly increase the quality of generated shapes compared to $10$k steps. We empirically observed that a smaller batch size improves results and noticed an optimum for the learning rate \num{1e-5}. For all our experiments, we use the best-performing hyperparameter setting.\\ 
The approach we adopted to choose between the hyperparameters was by generating 100 chairs with every model and then computing the following metrics which we have described in the main paper: MMD, COV and 1-NNA between the generated shapes and a reference set from the original data. The model scoring the lowest MMD and 1-NNA, and the highest COV values is the model which hyperparameters will be the reference for training our \name models.

\end{document}